%% file: acl2021.tex
\pdfoutput=1

\documentclass[11pt,a4paper]{article}
\usepackage{authblk}
\usepackage[hyperref]{acl2021}
\usepackage{latexsym}

\usepackage{hyperref}
\usepackage{amsmath}
\usepackage{amsfonts}
\usepackage{amssymb}
\usepackage{amsthm}
\usepackage{amsmath}
\usepackage{multirow}
\usepackage{bbm}
\usepackage{microtype}
\usepackage{graphicx}
\usepackage{booktabs} 
\usepackage[algo2e,vlined,linesnumbered,ruled,resetcount]{algorithm2e}
\usepackage{pgfplots}
\usepackage{subcaption}
\usepackage{tikz}
\usepackage{adjustbox}

\usepackage[utf8]{inputenc}
\usepackage{fourier} 
\usepackage{array}
\usepackage{makecell}

\usepackage{amsmath}
\newcommand\ltwonorm[1]{\left\lVert#1\right\rVert_{2}}
\newcommand\lonenorm[1]{\left\lVert#1\right\rVert_{1}}

\usepackage{microtype}

\aclfinalcopy 
\def\aclpaperid{***} 

\newcommand\blfootnote[1]{%
  \begingroup
  \renewcommand\thefootnote{}\footnote{#1}%
  \addtocounter{footnote}{-1}%
  \endgroup
}

\title{Direction is what you need: Improving Word Embedding Compression in Large Language Models}

\makeatletter
\newcommand\email[2][]%
   {\newaffiltrue\let\AB@blk@and\AB@pand
      \if\relax#1\relax\def\AB@note{\AB@thenote}\else\def\AB@note{\relax}%
        \setcounter{Maxaffil}{0}\fi
      \begingroup
        \let\protect\@unexpandable@protect
        \def\thanks{\protect\thanks}\def\footnote{\protect\footnote}%
        \@temptokena=\expandafter{\AB@authors}%
        {\def\\{\protect\\\protect\Affilfont}\xdef\AB@temp{#2}}%
         \xdef\AB@authors{\the\@temptokena\AB@las\AB@au@str
         \protect\\[\affilsep]\protect\Affilfont\AB@temp}%
         \gdef\AB@las{}\gdef\AB@au@str{}%
        {\def\\{, \ignorespaces}\xdef\AB@temp{#2}}%
        \@temptokena=\expandafter{\AB@affillist}%
        \xdef\AB@affillist{\the\@temptokena \AB@affilsep
          \AB@affilnote{}\protect\Affilfont\AB@temp}%
      \endgroup
       \let\AB@affilsep\AB@affilsepx
}
\makeatother

\author[$\dagger$*]{Klaudia Bałazy}
\author[$\ddagger$*]{Mohammadreza Banaei}
\author[$\ddagger$]{Rémi Lebret} 
\author[$\dagger$]{Jacek Tabor}
\author[$\ddagger$]{Karl Aberer}
\affil[$\dagger$]{Jagiellonian University} 
\email{\nolinkurl{klaudia.balazy@doctoral.uj.edu.pl, jacek.tabor@uj.edu.pl}}
\affil[$\ddagger$]{EPFL}
\email{\nolinkurl{[mohammadreza.banaei, remi.lebret, karl.aberer]@epfl.ch}}

\date{}

\usepackage{times}

\begin{document}
\maketitle

\blfootnote{$^{*}$Equal contribution}

\input{abstract.tex}
\input{introduction.tex}

\input{relatedworks.tex}
\input{model.tex}
\input{evaluation.tex}
\input{conclusion.tex}
\bibliographystyle{acl_natbib}
\bibliography{anthology,acl2021}


\end{document}

%% file: abstract.tex

\begin{abstract}
   The adoption of Transformer-based models in natural language processing (NLP) has led to great success using a massive number of parameters. However, due to deployment constraints in edge devices, there has been a rising interest in the compression of these models to improve their inference time and memory footprint. This paper presents a novel loss objective to compress token embeddings in the Transformer-based models by leveraging an AutoEncoder architecture. More specifically, we emphasize the~importance of the~direction of compressed embeddings with respect to original uncompressed embeddings. The proposed method is task-agnostic and does not require further language modeling pre-training. Our method significantly outperforms the commonly used SVD-based matrix-factorization approach in terms of initial language model Perplexity. Moreover, we evaluate our proposed approach over SQuAD~v1.1 dataset and several downstream tasks from the GLUE benchmark, where we also outperform the baseline in most scenarios. Our code is public.\footnote{\url{https://github.com/MohammadrezaBanaei/orientation_based_embedding_compression}}.
\end{abstract}

%% file: introduction.tex
\section{Introduction}
\label{sec:intro}

\begin{figure}[t!]
    \centering
    \includegraphics[width=0.3\textwidth,height=0.3\textwidth]{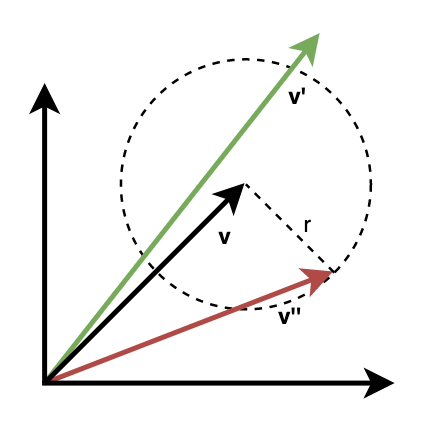}
    \caption{This figure presents a two-dimensional visualization of a token embedding vector $v$ with its two approximations: $v'$ and $v''$. Vector $v'$ has a larger Euclidean distance error than $v''$, but its direction is more similar to the reference vector. Our experiments show that $v'$ generally provides a better approximation of the original token compared to $v''$.}
    \label{fig:intuition}
\end{figure} 

Pretraining deep Transformer models \cite{vaswani2017attention} with language modeling and fine-tuning these models over downstream tasks have led to great success in recent years \cite{devlin2018bert, liu2019roberta, yang2019xlnet}, and even enabled researchers to design models that outperform human baselines in the GLUE benchmark \cite{wang2018glue}. Although these models are empirically powerful in many natural language understanding (NLU) tasks, they often require a massive number of parameters, making them hard to use for memory-constrained applications (e.g., edge devices). Therefore, there have been efforts to compress BERT-like models while preserving comparable performance with the original model.

Many of these compression methods are based on knowledge distillation \cite{hinton2015distilling} to help the compressed model (student) to perform close to the original model in different NLU tasks. However, these approaches often need high computation resources due to e.g., the necessity of retraining the expensive language modeling on a huge corpus \cite{sanh2019distilbert} or the use of expensive augmentation techniques to make the distillation effectively work \cite{jiao2019tinybert}. Moreover, compression techniques that rely on training/fine-tuning language models are becoming less feasible due to its ever-increasing cost for current state-of-the-art architectures with hundreds of millions of parameters \cite{he2020deberta, raffel2019exploring, brown2020language}.

More recently, there have been efforts to compress Transformer-based models for more resource-constrained scenarios \cite{mao2020ladabert} by using offline methods, such as matrix factorization \cite{winata2019effectiveness, lan2019albert, wang2019structured}, weight pruning \cite{li2016pruning, han2015learning}, and also weight quantization \cite{zhou2016dorefa, hubara2016binarized}.

This paper focuses on token embedding matrix compression due to being one of the largest matrices in BERT-based architectures. We specifically question the effectiveness of current low-rank matrix factorization methods in recent literature \cite{lan2019albert, wang2019structured} by comparing them with the performance of a linear AutoEncoder over different compression ratios\footnote{Number of parameters in the original embedding matrix, over the sum of the parameters in factorized matrices.}. We define a new loss objective which is not only dependent on the commonly used Mean Absolute Error (MAE) or Mean Squared Error (MSE) loss between input embeddings and AutoEncoder reconstruction, but is also sensitive to the noise in reconstructed embeddings "direction" (measured by cosine distance). We present the intuition behind the importance of embedding vector direction in the Figure \ref{fig:intuition}. In the following sections we show that cosine distance indeed plays a more critical role than MAE/MSE (Figure \ref{plot:crcos}) as measured by the Perplexity of the entire model in language modeling. 

In Section \ref{sec:result}, we demonstrate that our compression algorithm is superior or competitive to the Singular Value Decomposition (SVD) baseline over several natural language understanding tasks from GLUE \cite{wang2018glue} benchmark, as well as the SQuAD dataset \cite{rajpurkar2016squad} for question answering. We also compare our performance with the SVD-based compression over different compression ratios, and specifically show that our model performs consistently better in higher compression ratios.


Our contribution can be summarized as follows: 
\begin{itemize}
      \item We demonstrate the importance of direction (measured by cosine distance) in token embeddings compression.
      \item We leverage the AutoEncoder architecture to explore various multi-objective optimization functions.
      \item We outperform the SVD-based baseline in terms of Perplexity and over various downstream tasks.
 \end{itemize}


%% file: relatedworks.tex
\section{Related work}
\label{sec:relatedwork}
The current mostly used compression methods can be roughly categorized into four classes, namely knowledge distillation \cite{hinton2015distilling}, weight pruning \cite{li2016pruning, han2015learning}, matrix factorization \cite{lan2019albert, wang2019structured, mao2020ladabert} and weight quantization \cite{zhou2016dorefa, hubara2016binarized}. This section focuses on matrix factorization-based methods that are currently used for token embedding compression in the literature.

\subsection{Background: Low-rank matrix factorization}
This section describes the baseline method that we are comparing our approach with throughout the paper. Let $A$ be $n \times m$ embedding matrix representing $m$-dimensional embedding for each $n$ different input tokens. 
The truncated version of the matrix factorization aims to find a low-rank approximation $\widetilde{A}$ of input matrix $A$ \cite{halko2010finding}:
\begin{equation}
    \widetilde{A} = BC,
\end{equation}
where B is the size of $n \times k$ and C is the size of $k \times m$. When the inner dimension $k$ is smaller than $min(n, m)$, then the approximation is less expensive for storing it and performing further computations. The objective of this approximation is:

\begin{equation}
    L_2(A, \widetilde{A}) = \ltwonorm{A - \widetilde{A}},
\end{equation}
where $\ltwonorm{\cdot}$ denotes the $l_2$ operator norm. In this paper, we use the SVD method as a low-rank matrix factorization baseline to compare our approach.

\subsection{Matrix factorization for token embeddings compression}
\citet{lan2019albert} proposed to use matrix factorization to limit the number of parameters in the token embedding matrix, which also separates the Transformer hidden layer dimension from the size of vocabulary embedding. It is especially important as token embeddings are supposed to be \emph{context-independent}, but hidden layer representation should be a \emph{context-dependent} representation and hence needs more parameters. Moreover, reducing the vocabulary embedding dimension reduces the chance of overfitting, as many of the tokens are rarely used in downstream tasks.

There have been more recent efforts that use matrix factorization idea to compress different matrices in the Transformer architecture \cite{wang2019structured, mao2020ladabert}. For instance, \citet{mao2020ladabert} proposed an iterative hybrid approach that uses matrix factorization together with weight pruning (while distilling knowledge from a teacher model) until reaching the final desired compression ratio.  \citet{lioutas2019distilled} also proposed using a non-linear AutoEncoder model with knowledge distillation to compress word embeddings. However, we later demonstrate that only adding non-linearity indeed results in a minor improvement to the resulting compressed language model quality.

In this paper, we specifically focus on the effectiveness of SVD for compression of the token embedding matrix and show that Root Mean Square Error (RMSE) is not an optimal function to minimize the zero-shot Perplexity of the language model, which is the main criterion when language models are trained. We propose a new loss objective for linear matrix factorization using AutoEncoder to achieve a task-agnostic compressed language model with reasonable Perplexity without further fine-tuning the language model. In this work, we mainly investigate the effectiveness of SVD, and other complementary methods such as knowledge distillation can be used later to further boost the performance.


%% file: model.tex
\section{Model Description}

\begin{figure*}[h]
    \centering
    \includegraphics[width=0.75\textwidth,height=0.7\textwidth]{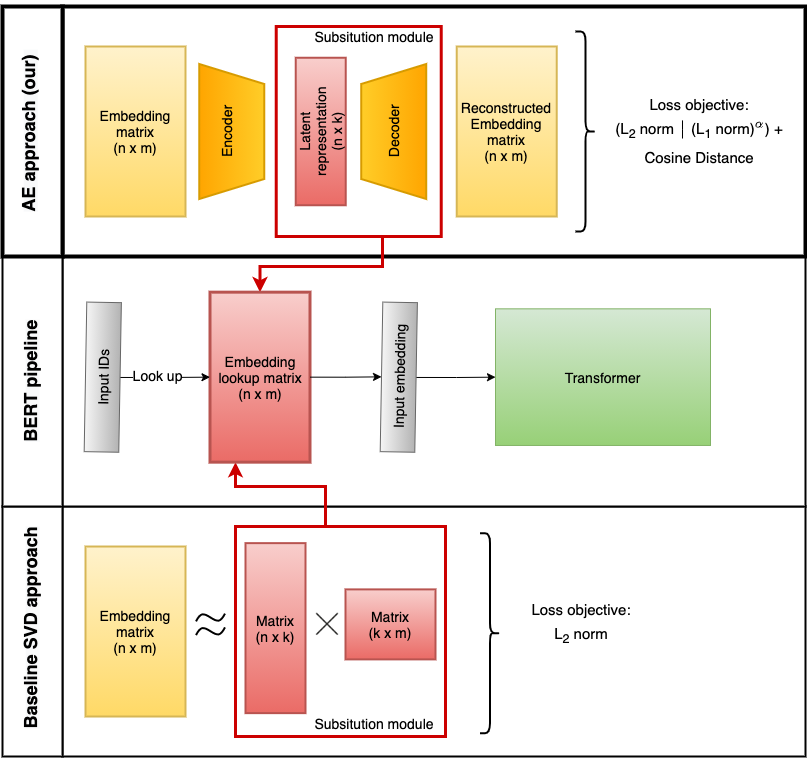}
    \caption{Overview of our AutoEncoder (ours) approach for BERT-like embedding matrix compression.}
    \label{fig:overview}
\end{figure*} 

Although SVD matrix-factorization is one of the most popular methods for matrix compression, we believe it is not an optimal method for compressing token embeddings in BERT-like architectures. The objective of SVD is to minimize the $l_2$~norm between the original matrix and the reconstructed one; however, focusing on $l_2$~norm optimization prioritizes the reduction of larger errors, and it may end up ignoring more minor vector differences. It is also sensitive to the influence of outliers. The most crucial reason for the $l_2$~norm not being the best choice is that it only considers the distance between the original and reconstructed token vector, and it does not necessarily pay attention to the orientation difference between them. In section \ref{sec:result}, we demonstrate that vectors representing language tokens are more sensitive to noise in their direction rather than to changes in Euclidean distance from the reference vector. We also discuss the motivation behind it further in this section.

In order to mitigate the problem of focusing only on the largest errors between two vectors, we propose replacing the $l_2$~norm objective with the $l_1$~norm raised to the power of $\alpha$: 
\begin{equation}
    \label{eq:alphal1}
    L_1^{\alpha}(A, \widetilde{A}) = \lonenorm{A - \widetilde{A}}^{\alpha},
\end{equation}
where $A$~denotes the original embedding matrix, $\widetilde{A}$~denotes the reconstructed embedding matrix, and $\lonenorm{\cdot}$~denotes the $l_1$~operator norm. Due to the flexibility in our defined loss objective, by decreasing the $\alpha$~parameter, we can control how much we want to focus on smaller error differences. We may set the $\alpha$~parameter to be a constant value, or linearly decrease it during the training. We denote linearly decreasing strategy for~$\alpha$~as:
\begin{equation}
    \label{eq:alphadec}
    [t_1, t_2],
\end{equation}
where $t_1$~is a starting value of $\alpha$~and $t_2$~is the target value to be reached at the end. The intuition behind using a decreasing $\alpha$~is to sequentially make the reconstruction harder for the model during training (as when the $\alpha$ becomes smaller, small reconstruction errors will also be magnified).

Since we believe that enforcing direction similarity between the original and the reconstructed embedding vectors is crucial for better language model performance, we introduce the second loss objective component, namely, cosine distance. Cosine distance can be interpreted as a measure of the difference in orientation of two vectors. This measure has been widely used in NLP for finding similar words~\cite{mikolov2013efficient}, document clustering~\cite{muflikhah2009document}, detecting plagiarism~\cite{foltynek2019academic}, and many more. The goal of introducing cosine distance loss as a part of our objective is to enforce direction similarity of each pair of vectors from the original and reconstructed matrix.

Taking into consideration all points above, we propose to replace the $l_2$~norm objective with a new multi-objective function consisting of $l_1$~norm (raised to the power of~$\alpha$, where $\alpha$~is a hyper-parameter that can be changed during training) and cosine distance:
\begin{equation}
    \label{eq:ourloss}
    \Phi_{\alpha, \beta}(A, \widetilde{A})  = L_1^{\alpha}(A, \widetilde{A}) + \beta * CD(A, \widetilde{A}),
\end{equation}
where $A$~denotes the original embedding matrix, $\widetilde{A}$~denotes the reconstructed embedding matrix, and $CD(A, \widetilde{A})$~represents the mean cosine distance of all embedding vector pairs. It is worth noting that it is the combination of these two functions that gives a powerful tool which allows both to optimize the distance and direction of the reconstructed vectors to the reference. Focusing only on one of these functions may lead to suboptimal results. For comparison, we also define another multi-objective function which is the combination of $l_2$~norm with cosine distance loss:
\begin{equation}
    \label{eq:l2cosloss}
    \Psi_{\beta}(A, \widetilde{A})  = L_2(A, \widetilde{A}) + \beta * CD(A, \widetilde{A}).
\end{equation}
In addition to the new loss function, we propose leveraging Auto-Encoder architecture for $\Phi_{\alpha, \beta}$~and $\Psi_{\beta}$~loss optimization~(Equation~\ref{eq:ourloss}~and~\ref{eq:l2cosloss}). We use a simple AutoEncoder consisting of a~one-layer Encoder/Decoder without any activation function in order to have a~fair comparison with the~SVD baseline. Using Auto-Encoder enables efficient multi-objective optimization, but it also allows to select the~appropriate level of model complexity when needed. At the end of the~Auto-Encoder training, we extract an~approximation of the original matrix, as shown in Figure~\ref{fig:overview}. We substitute the original embedding matrix with a new module consisting of latent representation of vocabulary tokens along with the Decoder module. 
\label{sec:model}

%% file: evaluation.tex
\section{Results}
\label{sec:result}

\input{cos_ae_svd}

In this section, we evaluate our approach, which is based on using AutoEncoder model with a multi-objective loss function that incorporates cosine distance with $l_1$~or~$l_2$~norm (Equation~\ref{eq:ourloss} and Equation~\ref{eq:l2cosloss}) on the task of BERT-like token embedding matrix compression. We compare our results versus the commonly used randomized SVD method \cite{halko2010finding} to perform low-rank matrix factorization. We have implemented our token embeddings compression with the PyTorch backend~\cite{paszke2019pytorch} and as an extension of Huggingface's Transformers library~\cite{wolf2019huggingface}, enabling researchers to apply our compression method in most of the existing Transformer architectures. It is worth noting that the offline training of our compression method on BERT-base~\cite{devlin2018bert} token embedding matrix takes only few minutes on a single GPU device.

\subsection{Experiments}

In this paper, we perform our experiments over BERT-base model, but the general idea can be applied to the vocabulary embeddings of any other similar transformer-based architecture. The BERT-Base token embedding matrix consists of more than 23 Million parameters which is around~21\% of all parameters in the model. We evaluate the~quality of our final compressed embeddings on the masked \cite{devlin2018bert} language modeling task (using WikiText-103 test dataset), GLUE benchmark \cite{wang2018glue} downstream tasks and SQuAD~v1.1 dataset \cite{rajpurkar2016squad}. We also analyze results on other metrics, namely RMSE, MAE and Cosine Distance.

In Figure~\ref{plot:crcos}, we compare the Perplexity score achieved by SVD\footnote{For SVD training, we select an iteration that minimizes Perplexity over our language modeling dataset.} method versus the results achieved by a linear AutoEncoder model with different loss configurations, when compressing BERT token embeddings. We specifically examine the importance of cosine distance coefficient~($\beta$) in our studied loss functions over three different compression ratios:~2.5, 5, 10. The loss objective~$\Phi_{t, \beta}$~ (Equation~\ref{eq:ourloss}) denotes constant (during the entire training) $\alpha$~parameter (equals to $t$) and~$\Phi_{[t_1,t_2], \beta}$~denotes linearly decreasing~$\alpha$~parameter (from~$t_1$~to~$t_2$). We present results when~$\alpha=0$, which represents combination of $l_1$~norm with cosine distance, and also when~$\alpha$ linearly decreases from 1.0 or from 2.0 to 0.6 ($[1.0, 0.6]$ and $[2.0, 0.6]$ respectively). These values have been selected experimentally.

\input{result_table}

\input{glue_table}

\input{glue_squad_results}

Table \ref{tab:aeresults} presents more metrics to compare SVD method with our AutoEncoder-based approach. We show the results of the model with the best performing objective function (in terms of Perplexity) for a given compression ratio. Additionally, we examine the effect of adding non-linear activation function to this selected AutoEncoder model, where it can be seen that the improvements due to addition of non-linearity is marginal.

We further validate the quality of our compressed token embeddings by inserting it into the BERT-base architecture and fine-tuning the model on different downstream tasks from the GLUE benchmark~\cite{wang2018glue} and on the SQuAD v1.1~\cite{rajpurkar2016squad} dataset. Table~\ref{tab:GLUE_results} presents an extensive comparison between our best (in terms of perplexity) linear AE and the SVD baseline on eight different downstream tasks and over different compression ratios. More specifically, we can see that our proposed method is superior or competitive to the SVD baseline and performs relatively better (compared to baseline) on higher compression ratios. The original BERT (without compression) performance is also added for a better comparison of studied scenarios.

Figure \ref{plot:squad} presents learning curves for three selected NLU downstream tasks: SST-2~\cite{socher2013recursive}, MRPC~\cite{dolan2005automatically} and SQuAD 1.1~\cite{rajpurkar2016squad}. We show results for the compression ratio of 10, as we observed more significant gain for higher compression ratios.

\subsection{Discussion}

The experiments presented in Figure~\ref{plot:crcos} confirm our claim that the $l_2$~norm alone is not an~optimal measure for evaluating the quality of reconstructed token embeddings in a Transformer-based architecture. We observe that adding cosine distance objective function correlates positively with a better Perplexity metric (Figure~\ref{plot:crcos}) and also with higher performance on downstream tasks~(Table~\ref{tab:GLUE_results}). Figure~\ref{plot:crcos} demonstrates that the best results are achieved when the~cosine distance coefficient $\beta$~is a~dominant component of~the loss function. However, if the~$\beta$~factor becomes too large, the quality of the solution decreases. Hence, we conclude that taking into account both the commonly used~L1/L2 distance and focusing on the direction of the token vectors are indispensable. We show that combining the $l_2$ or $l_1$ norm with the cosine distance into one multi-objective loss function and optimizing it by AutoEncoder model outperforms the baseline SVD Perplexity for all tested compression ratios~(Figure~\ref{plot:crcos}). Our experiments show that depending on the compression ratio $l_2$ or $l_1$ norm may be a better choice. However, they are conclusive that adding cosine distance is the key factor.

Moreover, our approach outperforms SVD in terms of accuracy for most GLUE benchmark downstream tasks and on SQuAD~v1.1~(Table~\ref{tab:GLUE_results}). We also observe that for higher compression ratios, our approach outperforms the SVD approach more significantly. More importantly, Figure~\ref{plot:squad} demonstrates that using our linear AutoEncoder compressed module in the BERT model generally converges faster than SVD-based compressed module, which is especially important in few-shot learning scenarios.

Looking at the results presented in Table~\ref{tab:aeresults}, we may also reflect on the importance of preserving the token vector orientation and its effect on Perplexity. More specifically, the mean cosine distance measures for SVD and our approach are pretty close, but its effect on Perplexity metric is significant. Our approach indeed provides a compressed submodule with a much better (lower) Perplexity.

We also show that only adding a non-linear activation function to the studied AutoEncoder model has a little effect on improving Perplexity. Table~\ref{tab:aeresults} presents the~effect of modifying the original linear AutoEncoder architecture by adding ELU~\cite{ELUclevert2016fast} as this activation shows a better impact on Perplexity than other activations in our experiments. It can be seen that the improvements in Perplexity due to the addition of non-linearities are marginal (as previously observed by \citet{lioutas2019distilled} in a distillation-based approach for token embeddings compression). Hence, we focused only on the linear AutoEncoder in all our downstream tasks experiments.

%% file: cos_ae_svd.tex
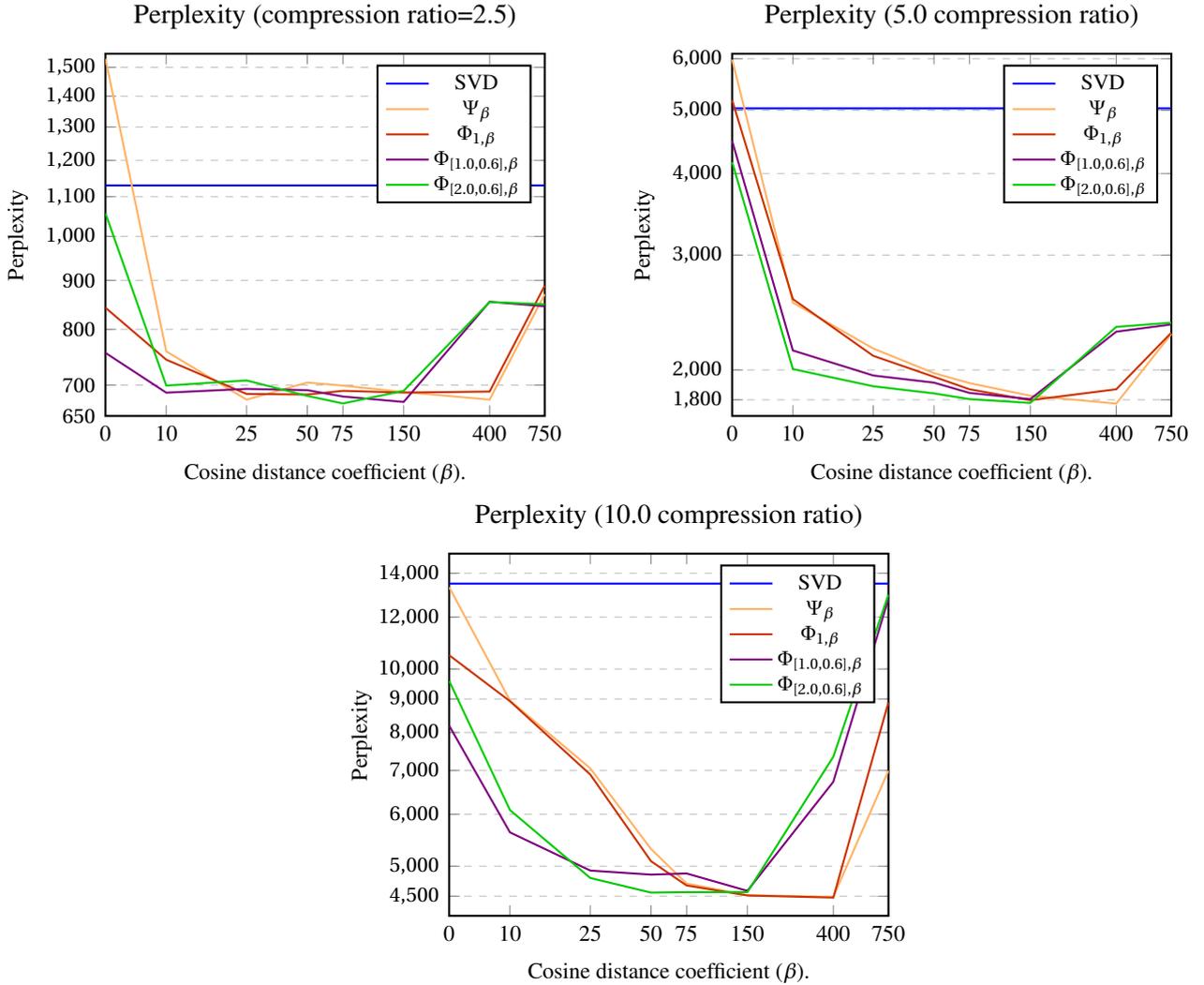
\begin{figure*}[tb!]
\centering
\begin{subfigure}[b]{0.45\textwidth}
\hspace*{-0.5cm} 
\begin{tikzpicture}
\begin{axis}[
    title={Perplexity (compression ratio=2.5)},
    xlabel={Cosine distance coefficient ($\beta$).},
    ylabel={Perplexity},
    xmin=5, xmax=750,
    ymin=650, ymax=1550,
    ymode=log,
    xmode=log,
    log ticks with fixed point,
    xtick={5,10,25,50,75,150,400,750},
    xticklabels={0,10,25,50,75,150,400,750},
    ytick={650, 700, 800, 900, 1000, 1100, 1200, 1300, 1400, 1500},
    label style={font=\small},
    tick label style={font=\small},
    legend pos=north east,
    legend style={nodes={scale=0.8, transform shape}},
    ymajorgrids=true,
    grid style=dashed,
    thick,scale=0.9, every node/.style={scale=0.9}
]
\addplot[
    color=blue,
    ]
    coordinates {
    (5,1130)(750,1130)
    };
    \addlegendentry{SVD}
\addplot[
    color=orange!60,
    ]
    coordinates {
    (5,1531)(10,759)(25,675.9)(50,704.3)(75,699.1)(150,688.4)(400,676)(750,870.5)
    };
    \addlegendentry{$\Psi_{\beta}$}
\addplot[
    color=red!80!green,
    ]
    coordinates {
    (5,842.7)(10,744)(25,685.3)(50, 684.3)(75,690.4)(150,687.5)(400,689.2)(750,888.5)
    };
    \addlegendentry{$\Phi_{1, \beta}$}
\addplot[
    color=violet,
    ]
    coordinates {
    (5,756.3)(10,687.3)(25,693.6)(50, 691.5)(75,681)(150,672.2)(400,854.8)(750,845.7)
    };
    \addlegendentry{$\Phi_{[1.0,0.6], \beta}$}
\addplot[
    color=black!20!green,
    ]
    coordinates {
    (5,1057)(10, 699.1)(25,707.8)(50, 682)(75,669.8)(150,690.3)(400,854.4)(750,849.8)
    };
    \addlegendentry{$\Phi_{[2.0,0.6], \beta}$}
\end{axis}
\end{tikzpicture}
\end{subfigure}%
\hfill
\begin{subfigure}[b]{0.45\textwidth}
\hspace*{-0.5cm} 
\begin{tikzpicture}
\begin{axis}[
    title={Perplexity (5.0 compression ratio)},
    xlabel={Cosine distance coefficient ($\beta$).},
    ylabel={Perplexity},
    xmin=5, xmax=750,
    ymin=1700, ymax=6106,
    ymode=log,
    xmode=log,
    log ticks with fixed point,
    xtick={5,10,25,50,75,150,400,750},
    xticklabels={0,10,25,50,75,150,400,750},
    ytick={1800, 2000, 3000, 4000, 5000, 6000},
    label style={font=\small},
    tick label style={font=\small},
    legend pos=north east,
    legend style={nodes={scale=0.8, transform shape}},
    ymajorgrids=true,
    grid style=dashed,
    thick,scale=0.9, every node/.style={scale=0.9}
]
\addplot[
    color=blue,
    ]
    coordinates {
    (5,5035)(750,5035) 
    };
    \addlegendentry{SVD}
\addplot[
    color=orange!60,
    ]
    coordinates {
    (5,5982)(10,2536)(25,2157)(50,1978)(75,1909)(150, 1826)(400,1776)(750,2271) 
    };
    \addlegendentry{$\Psi_{\beta}$}
\addplot[
    color=red!80!green,
    ]
    coordinates {
    (5,5183)(10,2568)(25,2102)(50, 1952)(75,1868)(150, 1797)(400, 1868)(750,2281) 
    };
    \addlegendentry{$\Phi_{1, \beta}$}
\addplot[
    color=violet,
    ]
    coordinates {
    (5,4488)(10,2142)(25,1961)(50, 1912)(75, 1844)(150,1804)(400,2289)(750,2348)
    };
    \addlegendentry{$\Phi_{[1.0,0.6], \beta}$}
\addplot[
    color=black!20!green,
    ]
    coordinates {
    (5,4164)(10, 2006)(25,1888)(50, 1841)(75,1804)(150,1780)(400,2328)(750,2363)
    };
    \addlegendentry{$\Phi_{[2.0,0.6], \beta}$}
\end{axis}
\end{tikzpicture}
\end{subfigure}
\hfill
\begin{subfigure}[b]{0.45\textwidth}
\begin{tikzpicture}
\begin{axis}[
    title={Perplexity (10.0 compression ratio)},
    xlabel={Cosine distance coefficient ($\beta$).},
    ylabel={Perplexity},
    xmin=5, xmax=750,
    ymin=4200, ymax=15000,
    ymode=log,
    xmode=log,
    log ticks with fixed point,
    xtick={5,10,25,50,75,150,400,750},
    xticklabels={0,10,25,50,75,150,400,750},
    ytick={4500, 5000, 6000, 7000, 8000, 9000, 10000, 12000, 14000},
    label style={font=\small},
    tick label style={font=\small},
    legend pos=north east,
    legend style={nodes={scale=0.8, transform shape}},
    ymajorgrids=true,
    grid style=dashed,
    thick,scale=0.9, every node/.style={scale=0.9}s
]
\addplot[
    color=blue,
    ]
    coordinates {
    (5,13501)(750,13501) 
    };
    \addlegendentry{SVD}
\addplot[
    color=orange!60,
    ]
    coordinates {
    (5,13330)(10,8949)(25,7052)(50,5311)(75,4705)(150,4516)(400,4489)(750,6995) 
    };
    \addlegendentry{$\Psi_{\beta}$}
\addplot[
    color=red!80!green,
    ]
    coordinates {
     (5,10496)(10,8930)(25,6900)(50,5088)(75,4673)(150,4511)(400,4478)(750,8896) 
    };
    \addlegendentry{$\Phi_{1, \beta}$}
\addplot[
    color=violet,
    ]
    coordinates {
    (5,8193)(10,5631)(25,4924)(50,4854)(75, 4874)(150, 4585)(400,6728)(750,12804) 
    };
    \addlegendentry{$\Phi_{[1.0,0.6], \beta}$}
\addplot[
    color=black!20!green,
    ]
    coordinates {
     (5,9589)(10,6090)(25,4797)(50, 4557)(75, 4564)(150, 4565)(400,7346)(750,12990)  
    };
    \addlegendentry{$\Phi_{[2.0,0.6], \beta}$}
\end{axis}
\end{tikzpicture}
\end{subfigure}
\caption{The impact of the $\protect\beta$ coefficient on Perplexity metric (lower is better) in the linear AutoEncoder loss functions: $\protect\Phi_{\alpha, \beta}$~(Equation~\ref{eq:ourloss}) and $\protect\Psi_{\beta}$~(Equation~\ref{eq:l2cosloss}). In all configurations we select a final model based on the best Perplexity achieved during training. The term~$\protect[t_1, t_2]$~indicates linearly decreasing $\protect\alpha$~parameter~(Equation~\ref{eq:alphadec}). Setting $\protect\beta = 0$ represents not including cosine distance component in the loss function. We may observe that not including cosine distance in the loss function as well as making it a too dominant component (very big $\protect\beta$) is not optimal for achieving good Perplexity. We also present the best Perplexity achieved by the baseline SVD method for three compression ratios: 2.5, 5.0, 10.0. Our approach significantly outperforms the baseline in the studied scenarios.} \label{plot:crcos} 
\end{figure*} 

%% file: result_table.tex
\begin{table*}[tb!]
    \small
    \centering
    \vskip 0.1in
    \begin{tabular}{@{}c@{\;}cccccccc}
         
         
         \midrule
         CR (\#Params) & Architecture & Objective & RMSE & Cosine Distance & MAE & Perplexity \\
         \midrule
         
        \multirow{2}{*}{\begin{tabular}[x]{@{}c@{}}2.5 (\textasciitilde9.38M) \end{tabular}} 
         & SVD & $l_2$
         & \textbf{0.02233} & 0.10300 & \textbf{0.01734} & 1130
         \\
         
         & Linear AE (+ ELU) & $\Phi_{[2.0,0.6], 75}$
         & 0.02427 (0.02431) & \textbf{0.1024 (0.1028)} & 0.01896 (0.01902) & \textbf{669.8 (664.0)}
         \\
         \midrule
         
         \multirow{2}{*}{\begin{tabular}[x]{@{}c@{}}5.0 (\textasciitilde4.69M)\end{tabular}} 
         & SVD & $l_2$
         & \textbf{0.02848} & 0.17490 & \textbf{0.02216} & 5035
         \\
         
         & Linear AE (+ ELU) & $\Psi_{400}$
         & 0.03101 (0.03061) & \textbf{0.17390 (0.17410)} & 0.02433 (0.02401) & \textbf{1776 (1730)}
         \\
         \midrule
         
         \multirow{2}{*}{\begin{tabular}[x]{@{}c@{}}10.0 (\textasciitilde2.34M)\end{tabular}} 
         & SVD & $l_2$
         & \textbf{0.03215} & 0.23050 & \textbf{0.02506} & 13501
         \\
         
         & Linear AE (+ ELU) & $\Phi_{1, 400}$
         & 0.03680 (0.03707) & \textbf{0.22900 (0.22910)} & 0.02909 (0.02934) & \textbf{4478 (4387)}
         \\
         \bottomrule
    \end{tabular}
    \caption{Additional metrics for comparing the performance of SVD baseline and the best performing linear AutoEncoder model (we select the configuration that minimizes Perplexity, as presented in Figure \ref{plot:crcos}) for different compression ratios (CR). For each AutoEncoder model, we also present (in parentheses) the results after adding non-linearity. Bold values indicate best results between SVD and linear AutoEncoder in each compression ratio.}
    \label{tab:aeresults}
\end{table*}

%% file: glue_table.tex
\begin{table*}[tb!]
    \small
    \centering
    \vskip 0.1in
    \begin{tabular}{@{}c@{\;}ccccccccc}
         
         
         \midrule
         CR & Architecture & \makecell{SST-2 \\ (Acc)} & \makecell{MRPC \\ (F1/Acc)} & \makecell{STS-B\\(Pearson/Spearman \\correlation)} & \makecell{QQP \\ (Acc/F1)} & \makecell{MNLI \\ (Acc)} & \makecell{QNLI \\ (Acc)} & \makecell{RTE \\ (Acc)} & \makecell{SQuAD v1.1 \\ (F1/EM)}\\
         \midrule
         
         - & Original BERT & 91.74 & 88.12/83.58 & 88.71/88.55 & 90.67/87.43 & 84.04 & 90.96 & 65.34 & 81.97/73.42  \\
         \midrule
         
        \multirow{2}{*}{\begin{tabular}[x]{@{}c@{}}2.5\end{tabular}}
         & SVD & 89.22 & 82.37/75.25 & 86.27/85.72 & 89.88/86.39 & 82.83 & \textbf{89.46} & \textbf{62.92} & 80.75/72.34
         \\
         
         & Linear AE & \textbf{90.83}
         & \textbf{86.64/80.88} & \textbf{87.35/86.88} & \textbf{90.04/86.72} & \textbf{83.13} & 89.16 & 62.58 & \textbf{81.29/72.85}
         \\
         \midrule
         
         \multirow{2}{*}{\begin{tabular}[x]{@{}c@{}}5.0\end{tabular}}
         & SVD & 87.04
         & 83.95/77.70 & 84.88/84.2 & \textbf{89.79/86.45} & \textbf{81.39} & 87.33 & 59.21 & 80.37/71.67
         \\
         
         & Linear AE & \textbf{88.07}
         & \textbf{86.67/81.37} & \textbf{85.9/85.43} & 89.2/85.66 & 81.11 & \textbf{87.53} & \textbf{64.26} & \textbf{80.53/72.00}
         \\
         \midrule
         
         \multirow{2}{*}{\begin{tabular}[x]{@{}c@{}}10.0\end{tabular}}
         & SVD & 82.0
         & 83.95/72.55 & 80.93/80.67 & 87.6/83.57 & 76.59 & 83.51 & 54.51 & 74.15/65.0
         \\
         
         & Linear AE & \textbf{84.29}
         & \textbf{84.06/77.7} & \textbf{84.7/84.16} & \textbf{88.32/84.38} & \textbf{79.26} & \textbf{86.09} & \textbf{58.48} & \textbf{75.70/66.75}
         \\
         \bottomrule
    \end{tabular}
    \caption{Performance comparison of the best SVD and the best linear AutoEncoder objective configuration on several NLU tasks from GLUE benchmark~\cite{wang2018glue} and for SQuAD v1.1 in different compression ratios (CR).}
    \label{tab:GLUE_results}
\end{table*}

%% file: glue_squad_results.tex
\pgfplotsset{compat = 1.3}

\begin{figure*}[tb!]
\centering
\begin{subfigure}[b]{0.33\textwidth}
\hspace*{-0.5cm} 
\begin{tikzpicture}
\begin{axis}[
    title={SST-2},
    xlabel={Epochs},
    ylabel={Accuracy},
    xmin=0.0, xmax=3,
    ymin=60, ymax=86,
    label style={font=\small},
    tick label style={font=\small},
    legend pos=south east,
    legend style={nodes={scale=0.7, transform shape}},
    ymajorgrids=true,
    grid style=dashed,
    thick,scale=0.6, every node/.style={scale=0.6}
]
\addplot[
    color=blue,
    ]
    coordinates {
     (0.0476, 60.67) (0.28571, 75.8) (0.57142, 79.1799) (0.8571, 79.47) (1.333, 81.34) (1.5238, 80.83) (1.71428, 81.17) (2.0, 82.2) (2.2857, 81.61) (2.5714, 81.73) (3.0, 81.94)
    };
    \addlegendentry{$SVD$}
\addplot[
    color=red!80!green,
    ]
    coordinates {
     (0.0476, 73.39) (0.28571, 77.18) (0.5714, 78.91) (0.85714, 81.0) (1.333, 82.24) (1.5238, 83.69) (1.7143, 84.42) (2.0, 83.78) (2.2857, 84.4) (2.5714, 84.22) (3.0, 84.28)
    }; 
    \addlegendentry{$AE$}
\end{axis}
\end{tikzpicture}
\end{subfigure}%
\hfill
\begin{subfigure}[b]{0.33\textwidth}
\begin{tikzpicture}
\begin{axis}[
    title={MRPC},
    xlabel={Epochs},
    ylabel={Accuracy},
    xmin=0.0, xmax=5,
    ymin=67, ymax=79,
    label style={font=\small},
    tick label style={font=\small},
    legend pos=south east,
    legend style={nodes={scale=0.7, transform shape}},
    ymajorgrids=true,
    grid style=dashed,
    thick,scale=0.6, every node/.style={scale=0.6}
]
\addplot[
    color=blue,
    ]
    coordinates {
     (0.30434, 68.38) (0.869, 69.86) (1.30434, 70.34) (1.7391, 71.65) (2.1739, 71.92) (2.6086, 72.6) (3.0434, 72.460) (3.4782, 72.72) (3.9130, 72.199) (4.34782, 72.3) (5.0, 72.53)
    };
    \addlegendentry{$SVD$}
\addplot[
    color=red!80!green,
    ]
    coordinates {
     (0.3043, 69.44) (0.8695, 73.07) (1.3043, 75.94)  (2.1739, 76.94) (2.608, 76.7) (3.0434, 77.72) (3.47826, 77.39) (3.913, 77.65) (4.3478, 77.7) (5.0, 77.67)
    }; 
    \addlegendentry{$AE$}
\end{axis}
\end{tikzpicture}
\end{subfigure}
\hfill
\begin{subfigure}[b]{0.33\textwidth}
\begin{tikzpicture}
\begin{axis}[
    title={SQuAD v1.1},
    xlabel={Epochs},
    ylabel={F1},
    xmin=0.0, xmax=3.0,
    ymin=30.0, ymax=80.0,
    label style={font=\small},
    tick label style={font=\small},
    legend pos=south east,
    legend style={nodes={scale=0.7, transform shape}},
    ymajorgrids=true,
    grid style=dashed,
    thick,scale=0.6, every node/.style={scale=0.6}
]
\addplot[
    color=blue,
    ]
    coordinates {
     (0.09,30.99)(0.18,40.09)(0.27,47.17)(0.36,52.53)(0.45,57.28)(0.54,58.36)(0.63,60.14)(0.72,62.32)(0.81,63.71)(0.9,64.98)(1.0, 65.86)(1.25, 68.34)(1.5, 69.19)(1.75, 71.15)(2.0, 72.36)(2.25, 72.85)(2.5, 73.27)(2.75, 73.63)(3.0, 74.15)
    };
    \addlegendentry{$SVD$}
\addplot[
    color=red!80!green,
    ]
    coordinates {
     (0.09,42.67)(0.18,49.48)(0.27,56.01)(0.36,58.42)(0.45,61.39)(0.54,63.71)(0.63,64.08)(0.72,66.38)(0.81,68.11)(0.9,68.89)(1.0, 69.16)(1.25, 71.39)(1.5, 72.24)(1.75, 73.6)(2.0, 74.16)(2.25, 74.49)(2.5, 74.82)(2.75, 75.34)(3.0, 75.7)
    }; 
    \addlegendentry{$AE$}
\end{axis}
\end{tikzpicture}
\end{subfigure}
\caption{Comparing the learning curves of the best SVD baseline and the best-selected configuration of the AutoEncoder model for SST-2 \cite{socher2013recursive}, MRPC \cite{dolan2005automatically}, and SQuAD v1.1 \cite{rajpurkar2016squad} during fine-tuning for compression ratio=10.0 .} \label{plot:squad} 
\end{figure*}
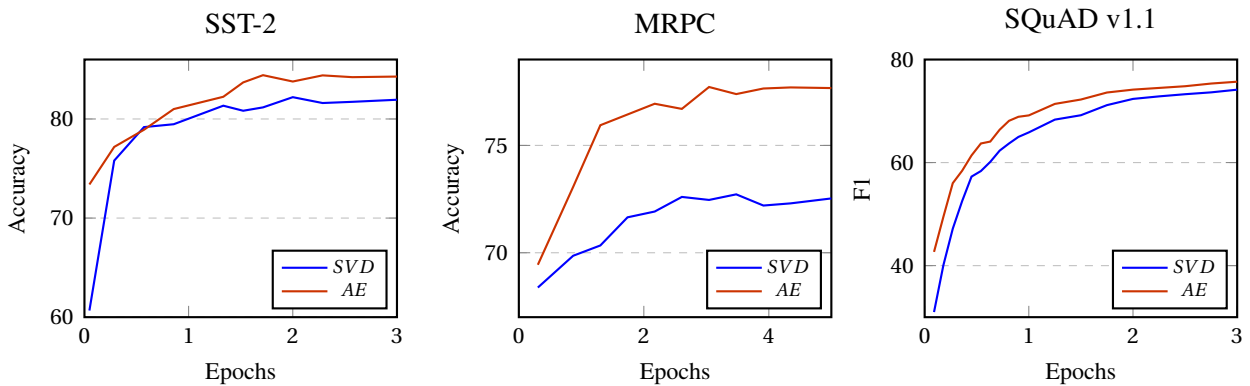 

%% file: conclusion.tex
\section{Conclusion}
In this work, we propose a simple linear AutoEncoder model with a multi-objective loss function for BERT-like token embeddings compression. We emphasize the importance of the direction component (measured by the cosine distance between the original and the reconstructed token embeddings) in the compression objective function. We challenge the commonly used SVD-based matrix-factorization method and show that our approach achieves significantly better zero-shot language model Perplexity. Moreover, we show that BERT-like models with our compressed token embeddings submodule converge much faster and outperform the SVD baseline on SQuAD~v1.1 and on GLUE benchmark tasks in most scenarios.


\section{Acknowledgements}
This research was partially funded by the Priority Research Area Digiworld under the program Excellence Initiative – Research University at the Jagiellonian University in Kraków.